\documentclass[lettersize,journal]{IEEEtran}
\usepackage{amsmath,amsfonts}
\usepackage{algorithm}
\usepackage{array}
\usepackage[caption=false,font=normalsize,labelfont=sf,textfont=sf]{subfig}
\usepackage{textcomp}
\usepackage{stfloats}
\usepackage{url}
\usepackage{verbatim}
\usepackage{graphicx}
\usepackage{cite}
\usepackage{multirow}
\usepackage{booktabs}
\usepackage{algpseudocode}  
\floatname{algorithm}{Algorithm}  

\begin{document}

\title{SAGOnline: Segment Any Gaussians Online}

\author{Wentao Sun, Quanyun Wu, Hanqing Xu, Kyle Gao,~\IEEEmembership{Member,~IEEE}, Zhengsen Xu, Yiping Chen,~\IEEEmembership{Senior Member,~IEEE}, Dedong Zhang,~\IEEEmembership{Member,~IEEE},  Lingfei Ma,~\IEEEmembership{Member,~IEEE}, John S. Zelek,~\IEEEmembership{Member,~IEEE}, Jonathan Li,~\IEEEmembership{Fellow,~IEEE}%
\thanks{Corresponding author: Yiping Chen}
\thanks{W. Sun, Q. Wu, K. Gao, D. Zhang, J. S. Zelek, and J. Li are with the University of Waterloo, Waterloo, ON N2L 3G1, Canada (e-mails: w27sun@uwaterloo.ca; q34wu@uwaterloo.ca; y56gao@uwaterloo.ca; dedong.zhang@uwaterloo.ca; jzelek@uwaterloo.ca; junli@uwaterloo.ca).}%
\thanks{H. Xu and L. Ma are with East China Normal University, Shanghai 200050, China (e-mails: 51273901140@stu.ecnu.edu.cn; l53ma@cufe.edu.cn).}%
\thanks{Z. Xu is with the University of Calgary, Calgary Alberta T2N 1N4, Canada (e-mail: zhengsen.xu@ucalgary.ca)}
\thanks{Y. Chen is with Sun Yat-Sen University, Zhuhai 519082, China (e-mail: chenyp79@mail.sysu.edu.cn).}%
}



\maketitle

\begin{abstract}
  3D Gaussian Splatting has emerged as a powerful paradigm for explicit 3D scene representation, yet achieving efficient and consistent 3D segmentation remains challenging. Existing segmentation approaches typically rely on high-dimensional feature lifting, which causes costly optimization, implicit semantics, and task-specific constraints. We present \textbf{Segment Any Gaussians Online (SAGOnline)}, a unified, zero-shot framework that achieves real-time, cross-view consistent segmentation without scene-specific training. SAGOnline decouples the monolithic segmentation problem into lightweight sub-tasks. By integrating video foundation models (e.g., SAM 2), we first generate temporally consistent 2D masks across rendered views. Crucially, instead of learning continuous feature fields, we introduce a \textbf{Rasterization-aware Geometric Consensus} mechanism that leverages the traceability of the Gaussian rasterization pipeline. This allows us to deterministically map 2D predictions to explicit, discrete 3D primitive labels in real-time. This discrete representation eliminates the memory and computational burden of feature distillation, enabling instant inference. Extensive evaluations on NVOS and SPIn-NeRF benchmarks demonstrate that SAGOnline achieves state-of-the-art accuracy (92.7\% and 95.2\% mIoU) while operating at the fastest speed at 27 ms per frame. By providing a flexible interface for diverse foundation models, our framework supports instant prompt, instance, and semantic segmentation, paving the way for interactive 3D understanding in AR/VR and robotics.
\end{abstract}


\section{Introduction}
\label{sec:intro}
\IEEEPARstart{T}{he} recent emergence of 3D Gaussian Splatting (3DGS)~\cite{3dgs} has reshaped real-time 3D scene reconstruction by offering an explicit, compact, and efficient scene representation. Unlike earlier implicit radiance-field formulations, 3DGS models a scene as a set of anisotropic Gaussian primitives, enabling real-time rendering while preserving high-fidelity geometry and appearance. Because this representation is discrete and explicitly structured, each Gaussian can be enriched with additional attributes such as semantic labels or feature vectors, making 3DGS well suited for downstream 3D understanding.

This ability to attach semantics to individual Gaussians is especially appealing for tasks that require online and cross-view consistent segmentation. Such capabilities are increasingly crucial for applications in augmented and virtual reality~\cite{survey1,ar1,vr3}, robotic manipulation and mapping~\cite{vr2-slam,slam2,slam3}, and autonomous systems~\cite{auto1,auto2}, where agents must parse the environment from continuously changing viewpoints. Consequently, a growing body of work aims to embed semantic information into 3D Gaussians so that the representation itself supports view-consistent segmentation from novel viewpoints~\cite{ClickGaussians,feature3dgs,SAGAgaussian}.
\IEEEpubidadjcol

Recent advancements in 3DGS-based segmentation can be broadly categorized by their mechanism for semantic injection. Distillation-based methods transfer features from 2D foundation models into 3D primitives, enabling early demonstrations of zero-shot 3D segmentation once the features are aligned with the Gaussian representation~\cite{feature3dgs,omniseg}. Contrastive frameworks, conversely, take potentially inconsistent multi-view 2D masks as input and enforce cross-view consistency in the 3D embedding space, which in turn enables consistent novel-view segmentation~\cite{SAGAgaussian,gaussian_grouping}. Interaction-based approaches operate directly on Gaussians to support click- or prompt-driven tasks and benefit from the geometry-aware structure of the Gaussian scene~\cite{ClickGaussians}. 

Despite these advancements, online 3DGS segmentation is still difficult due to three main challenges: First, inconsistent 2D observations. In an online system, 2D foundation models often produce unstable results as the camera viewpoint changes. Maintaining a stable 3D identity without slow, per-scene optimization is a major challenge. Second, complex occlusions. 3DGS consists of many overlapping primitives. In crowded scenes, it is hard to map 2D masks to the correct 3D depth, especially when objects are partially hidden from certain angles. Third, diverse task requirements. Different applications need different types of segmentation, such as instance, semantic, or prompt-based masks. A single, unified 3D representation that handles all these tasks without retraining is still missing.

Existing methods only partially address these challenges and introduce their own bottlenecks. First, to fix 2D inconsistencies, many methods~\cite{feature3dgs,omniseg} require long optimization times (e.g., tens of minutes). This makes real-time use impossible. Second, most approaches rely on large feature vectors instead of simple labels~\cite{feature3dgs,SAGAgaussian}. These features use a lot of memory. More importantly, they are "implicit," meaning the 3D masks are not directly visible. This makes it hard to handle complex occlusions because the system cannot accurately map labels to the correct depth. Finally, current pipelines~\cite{gaussian_grouping} are often rigid and designed for only one task. They lack a unified way to handle instance, semantic, and prompt segmentation at the same time.

To address these challenges, we propose Segment Any Gaussians Online (SAGOnline), a decoupled and optimization-free framework. Since 3DGS can render novel viewpoints at real-time frame rates, we can treat the continuous rendering stream as a synchronized video sequence. This physical characteristic allows us to decompose the monolithic 3D segmentation task into three lightweight, modular sub-tasks: (1) novel-view rendering, (2) multi-view consistent 2D video segmentation, and (3) global 3D consensus fusion. By bridging the domain gap between 3D scenes and 2D videos, we can directly deploy powerful video segmenters without any 3D fine-tuning or per-scene optimization.

To address the consistency challenge (Challenge 1), we leverage the high-speed rendering nature of 3DGS by treating its continuous rendering stream as a synchronized video sequence. This perspective allows us to bridge the domain gap between static 3D scenes and dynamic 2D observations, enabling the integration of temporal priors from state-of-the-art video segmenters to stabilize multi-view observations. By decoupling the segmentation task into a video-like processing pipeline, SAGOnline effectively mitigates the 2D inconsistency problem without requiring expensive per-scene optimization.

To overcome the occlusion and mapping ambiguity (Challenge 2), we introduce a Rasterization-aware Geometric Consensus (RGC) mechanism. By exploiting the deterministic traceability inherent in the 3DGS rasterizer, RGC establishes a robust voting scheme that fuses 2D labels into explicit, discrete 3D attributes for each primitive. This mechanism naturally handles complex occlusions by selectively aggregating semantic information only from the visible surface of the objects during the fusion process. Unlike prior implicit feature-based methods that suffer from "black-box" decoding, our explicit representation ensures high-fidelity mask projection and rigorous geometric interpretability, even in cluttered or densely occluded environments.

Furthermore, to ensure cross-task generalization (challenge 3), we design a generalizable interface for integrating foundation models. Because SAGOnline cleanly decouples rendering, 2D segmentation, and 3D fusion, various foundation models (e.g., SAM2~\cite{sam2}, SAM3~\cite{sam3}, SEEM~\cite{SEEM}, and video segmenters) can be plugged in interchangeably as modular components. Unlike prior methods that treat foundation models merely as offline mask generators, SAGOnline incorporates them as interactive components, enabling prompt segmentation, semantic segmentation, and instance segmentation within a unified pipeline.

Quantitative evaluations on the NVOS and SPIn-NeRF benchmarks demonstrate that SAGOnline outperforms existing baselines (e.g., Feature3DGS~\cite{feature3dgs}, SA3D~\cite{sa3d}) with state-of-the-art mIoUs of 92.7\% and 95.2\%, respectively. Regarding speed, our framework achieves immediate usability with the fastest novel-view segmentation rate of 27 ms per frame ($960 \times 540$ resolution on an NVIDIA RTX 4090). Beyond standard benchmarks, we validate the method's versatility on the KITTI-360~\cite{kitti360}, UDD~\cite{UDDdataset}, etc. datasets, proving its efficacy across diverse scenarios, including autonomous driving, drone-based urban views, and complex indoor scenes. Our main contributions are summarized as follows:
\begin{itemize}
    \item \textbf{A Novel Optimization-free Online Segmentation Framework:} We propose SAGOnline, a decoupled architecture that reinterprets the continuous rendering stream of 3DGS as a synchronized video sequence. By leveraging temporal priors from video foundation models, SAGOnline achieves robust cross-view consistency and eliminates the need for expensive per-scene optimization, enabling immediate and real-time 3D segmentation.
    \item \textbf{The Rasterization-aware Geometric Consensus (RGC) Mechanism:} We introduce RGC, a deterministic fusion scheme that bridges the gap between 2D observations and 3D primitives. By exploiting the traceability of the 3DGS rasterizer, RGC maps 2D predictions into explicit, discrete 3D attributes. This mechanism effectively resolves occlusion-induced ambiguities and depth-mapping inaccuracies inherent in implicit feature-based representations.
    \item \textbf{A Unified and Extensible Interface for Multi-task Segmentation:} Through its modular design, SAGOnline provides a task-agnostic interface that seamlessly integrates various foundation models (e.g., SAM2, SEEM). This unified pipeline supports instance, semantic, and prompt-based segmentation within a single framework without retraining, demonstrating superior versatility across diverse indoor and outdoor scenarios.
    \item \textbf{State-of-the-Art Performance and Efficiency:} Quantitative evaluations on NVOS and SPIn-NeRF benchmarks show that SAGOnline achieves record-breaking mIoU scores while maintaining a high-speed inference rate of 27 ms per frame. Its robustness is further validated on large-scale datasets including KITTI-360 and UDD, underscoring its potential for practical deployment in autonomous systems.
\end{itemize}

\section{Related Work}
\label{sec;relatedwk}
\subsection{Geometry-Aware 3D Gaussian Representations} The fidelity of 3D segmentation is intrinsically linked to the quality of the underlying geometric representation. While the original 3D Gaussian Splatting (3DGS) excels in photorealistic rendering, its optimization often results in unstructured volumetric "clouds" or "floaters" that lack physical surface fidelity~\cite{Sugar}. Such artifacts are catastrophic for segmentation, as semantic boundaries cannot be clearly defined on fuzzy geometries.

To address this, recent research has pivoted towards geometry-aware optimization. SuGaR~\cite{Sugar} regularizes Gaussians to align with the zero-level set of a Signed Distance Function (SDF), enabling explicit mesh extraction. More significantly for segmentation, 2D Gaussian Splatting (2DGS)~\cite{2dgs} and its successors~\cite{refGS} replace 3D ellipsoids with 2D oriented disks. This "planar shift" provides explicit normal vectors and enforces surface constraints, which prevent semantic labels from bleeding through the object volume. Our work builds upon these geometric insights, leveraging surface-aligned primitives to ensure that segmentation masks respect physical boundaries.

\subsection{Semantic Distillation and Consistency} 
\label{sec:related_distillation} 
Bridging the gap between 2D foundation models (e.g., SAM \cite{sam}, CLIP\cite{clip}) and 3D space is primarily achieved through feature distillation. Early attempts like Feature3DGS~\cite{feature3dgs} lift 2D semantic features into high-dimensional Gaussian attributes. However, these methods suffer from severe multi-view inconsistency, where conflicting 2D predictions lead to fragmented 3D boundaries. Recent studies, such as Feature-Homogenized GS (FHGS)~\cite{FHGS}, attribute this issue to the conflict between the anisotropic nature of 3DGS (optimized for view-dependent radiance) and the isotropic nature of semantic identity (invariant to view).

To mitigate inconsistency, contrastive learning frameworks like OmniSeg3D-GS~\cite{omniseg} and GLS~\cite{gls} employ cross-view clustering or joint optimization with geometric priors (e.g., TSDF). While these methods improve coherence, they typically operate in an offline manner, requiring expensive pre-computation or global optimization (e.g., Linear Programming in FlashSplat~\cite{flashsplat}) that precludes real-time user interaction.

\subsection{Interactive and Online Segmentation} 
\label{sec:related_interaction} 
The transition from static annotation to interactive editing has driven the development of human-in-the-loop systems. Early interactive models like SA3D~\cite{sa3d} relied on iterative inverse rendering, resulting in high latency (seconds to minutes per prompt).

To achieve real-time performance, recent approaches such as SAGA~\cite{SAGAgaussian} and Click-Gaussian~\cite{ClickGaussians} pre-cache semantic features, allowing for millisecond-level inference. SAGA specifically introduces scale-aware embeddings to disentangle semantic granularity from geometric scale. However, these methods are often limited by the temporal inconsistency of the underlying 2D segmentation models (SAM 1.0). With the advent of video foundation models like SAM 2~\cite{sam2}, new paradigms such as Seg-Wild~\cite{segwild} and WildSeg3D~\cite{wildseg} have begun to exploit temporal memory for better consistency. Yet, effectively integrating these video-based priors into a truly online, geometry-aware 3DGS framework—without succumbing to "spiky" artifacts or requiring offline retraining—remains an open challenge.

\subsection{Summary} 
In summary, existing paradigms largely treat geometry, semantics, and interaction as separate optimization targets. \textit{Geometry-focused} methods (SuGaR, 2DGS) lack semantic flexibility; \textit{Distillation-based} methods (Feature3DGS) struggle with consistency; and \textit{Interactive} frameworks (SA3D, SAGA) often trade geometric precision for speed. Our proposed method addresses this tripartite gap by unifying geometry-aware regularization with an online, interaction-driven mechanism enabled by foundation video models.

\begin{figure*}[t]
    \centering
    \includegraphics[width=1.0\linewidth]{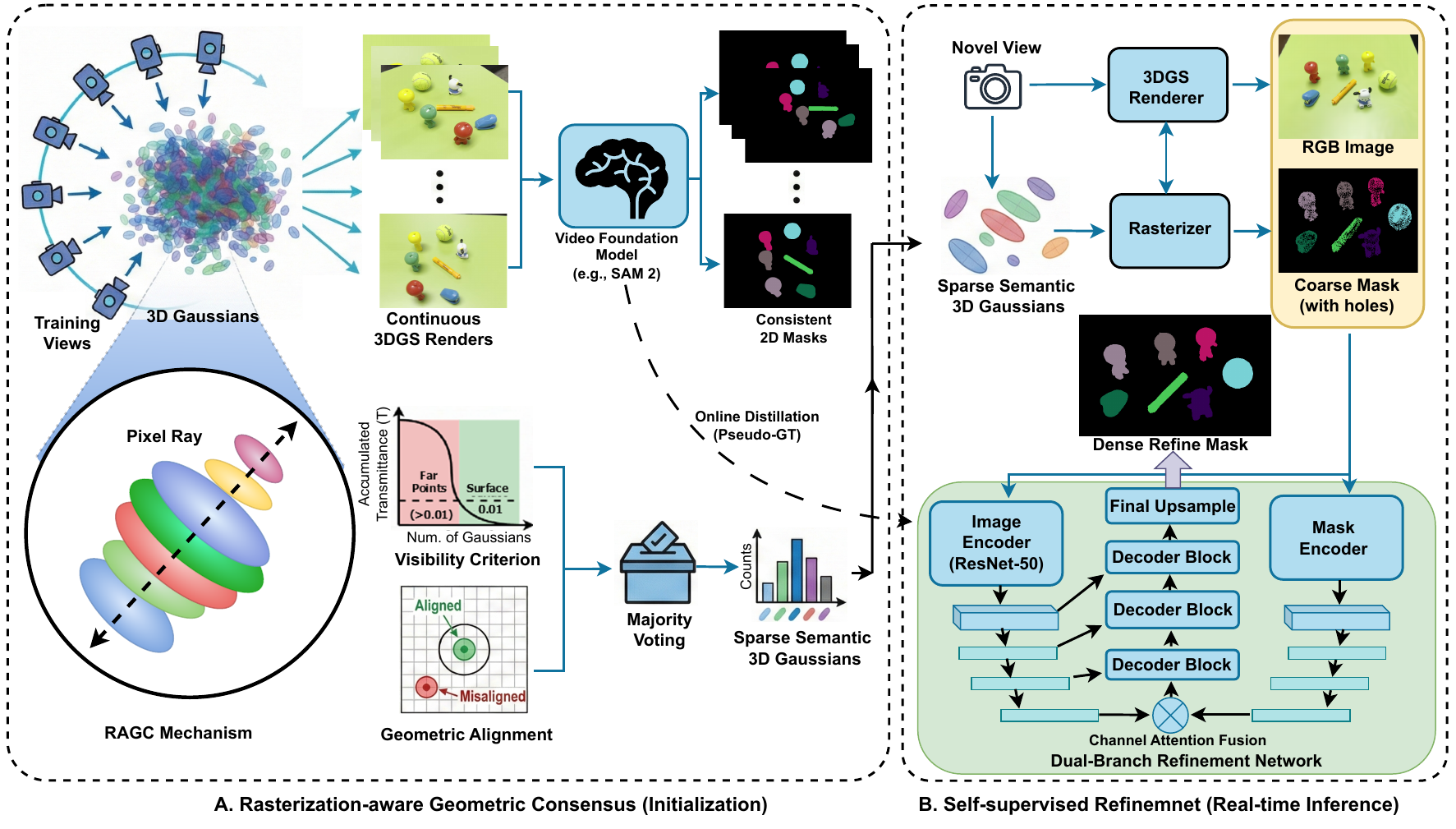}
    \caption{The architecture of SAGOnline. The framework comprises an initialization stage (Left) and a real-time inference stage (Right). (A) We propose a Rasterization-aware Geometric Consensus (RAGC) module to resolve semantic ambiguities. By analyzing the pixel ray, RAGC identifies valid Gaussians within the surface crust (green zone) that are geometrically aligned with the pixel center, aggregating 2D semantics via majority voting. (B) To achieve real-time segmentation, the learned sparse semantic Gaussians are projected to form a coarse mask. This coarse prior is fused with the photorealistic RGB render in a Dual-Branch Refinement Network, employing an encoder-decoder structure with channel attention to recover fine-grained details, supervised by online distillation.}
    \label{fig:pipeline}
\end{figure*}

\section{Method}
\label{sec:method}
The proposed SAGOnline framework aims to achieve real-time 3D semantic segmentation through two stages: Stage I: Rasterization-aware Geometric Consensus and Stage II: Self-supervised Refinement. Our key insight is that while foundational video segmenters provide consistent 2D masks, they are too heavy for real-time interaction. Therefore, we utilize them only as "teachers" to initialize 3D primitive labels and supervise a lightweight "student" network for instant inference.

\subsection{Problem Formulation}
Given a pre-trained Gaussian Splatting model (trained by 2DGS~\cite{2dgs}) $\mathcal{G} = \{g_i\}_{i=1}^N$, where each Gaussian $g_i$ is parameterized by its position $\mu$, rotation $q$, scale $s$, opacity $\alpha$, and color $c$. Our objective is twofold:
\begin{enumerate}
\item \textbf{3D Labeling:} Assign a discrete semantic label $l_i \in \{1, \dots, K\}$ to each primitive $g_i$, resulting in a semantic field $\mathcal{S} = \{l_i\}_{i=1}^N$.
\item \textbf{Real-time Segmentation:} Given an arbitrary camera pose $\theta$, efficiently render a high-quality semantic mask $\hat{M}_\theta \in \{1, \dots, K\}^{H \times W}$.
\end{enumerate}
The challenge lies in achieving this without high-dimensional feature lifting while maintaining temporal consistency and real-time inference speeds.

\subsection{Stage I: Rasterization-aware Geometric Consensus} 
This stage establishes a deterministic mapping between 2D multi-view semantics and 3D Gaussian primitives. By leveraging the inherent traceability of the 3DGS rasterization pipeline, we avoid the memory-intensive requirements of feature-based lifting. This module operates in two sequential steps: 2D temporal labeling and 3D geometric consensus.

\subsubsection{2D Temporal Labeling}
To generate reliable 2D supervision, we exploit the explicit nature of 3DGS, where rendering along a smooth camera trajectory naturally yields temporally coherent image sequences. Let $\mathcal{V} = \{v_1, v_2, \dots, v_T\}$ denote a sequence of camera poses. We first render the corresponding RGB images $\mathcal{I}$ using the 3DGS rasterizer:
\begin{equation}
\mathbf{I}_t = \mathcal{R}(\mathcal{G}, v_t),
\end{equation}
where $\mathcal{R}(\cdot)$ denotes the rasterization operator. To ensure cross-view consistency, we employ a foundational video segmenter $f_m$ (e.g., SAM 2 \cite{sam2}) to process the rendered sequence $\mathcal{I}$, producing a set of consistent 2D masks:
\begin{equation}
\mathbf{L}_{2D} = f_m(\mathbf{I}_1, \dots, \mathbf{I}_T).
\end{equation}
This video-based approach significantly mitigates the inconsistent issues common in frame-wise segmentation, providing a robust foundation for 3D label lifting.

\subsubsection{3D Geometric Consensus Mechanism}
Once the consistent 2D masks $\mathbf{L}_{2D}$ are obtained, we propose a \textit{Rasterization-aware Geometric Consensus} (RGC) mechanism to lift these labels into 3D space. During the 3DGS forward pass for a pixel $p$, the rendered color is computed via alpha-blending:
\begin{equation}
C(p) = \sum_{i \in \mathcal{N}p} c_i \alpha_i T_i, \quad 
T_i = \prod_{j=1}^{i-1} (1 - \alpha_j),
\end{equation}
where $\mathcal{N}_p$ denotes the sorted set of primitives overlapping pixel $p$, and $T_i$ represents the accumulated transmittance. To ensure that labels are only assigned to primitives that accurately represent the scene geometry, we enforce two strict filtering criteria:
\begin{itemize}
\item \textbf{Surface Visibility:} A primitive contributes to the semantic consensus only if it lies on the visible surface crust. We traverse the ray front-to-back and evaluate the \textit{post-primitive} transmittance $T_{i+1} = T_i (1-\alpha_i)$. A primitive $G_i$ is considered a valid surface component if the ray remains visibly potent upon reaching it ($T_i > \epsilon$). The traversal is terminated immediately when the accumulated opacity saturates, i.e., when $T_{i+1}$ drops below a threshold $\epsilon$ (experimentally set to $0.01$). This ensures we strictly label the visible surface layer while discarding fully occluded internal primitives.
\item \textbf{Geometric Alignment:} Due to the anisotropic nature of 3DGS, a primitive may intercept a ray with its elongated "tail" rather than its core, leading to loose semantic associations. To enforce strict geometric alignment, we require the projected center $\mu_i^{2D}$ of primitive $G_i$ to be strictly aligned with the pixel $p$: $\|\mu_i^{2D} - p\|_2 < \delta$ ($\delta$ experimentally set to $2$). This rejects primitives that only intersect the ray with their Gaussian tails, ensuring that the semantic label is assigned to the primitive's core.
\end{itemize}
Based on these criteria, we maintain a voting histogram $H_i$ for each Gaussian primitive $G_i$. For every pixel $p$ in the training views, if $G_i$ satisfies both the visibility and alignment criteria, it receives a vote from the corresponding 2D mask label $L_{2D}(p)$. The final 3D label $l_i$ is determined by the majority vote:
\begin{equation}
l_i = \underset{k}{\arg\max} \ H_i(k)
\end{equation}
Primitives that fail to accumulate sufficient votes or ambiguous votes are left unlabeled. This strict filtering results in a sparse but highly accurate explicit 3D segmentation, effectively creating a "semantic point cloud" on the object surface. This sparsity motivates the need for our subsequent self-supervised refinement stage.

\subsection{Stage II: Self-supervised Refinement}
\label{subsec:refinement}

Stage II is designed to operate concurrently with Stage I to overcome three critical limitations. First, the efficiency bottleneck: the segmentation speed in Stage I is strictly bounded by the video segmenter. Second, the viewpoint constraint: reliance on temporal continuity prevents the system from handling wide-baseline camera transitions. Third, the sparsity issue: although the \textit{Stage I: Rasterization-aware Geometric Consensus} effectively lifts 2D semantics to 3D, our rigorous filtering (to ensure precision) results in sparse 3D labels. Consequently, direct rasterization produces coarse masks ($M^{2D}_{\text{coarse}}$) with structural holes and aliasing. 

To achieve more rapid and robust Gaussian segmentation, we propose a Self-supervised Refinement Module. This module employs a lightweight student network, $\mathcal{N}_\phi$, to distill knowledge from two sources: the "accurate but slow" foundation model and the "consistent but sparse" 3D projections. By doing so, $\mathcal{N}_\phi$ transforms these inputs into a "fast and dense" inference engine, enabling high-quality segmentation across arbitrary viewpoints. The operational flow of this module is formulated as follows:
\begin{equation}
M^{2D}_{\text{coarse}} = \mathcal{R}(\mathcal{S}, v_t), \quad 
\mathbf{I}_t = \mathcal{R}(\mathcal{G}, v_t)
\end{equation}
\begin{equation}
M^{2D}_{\text{refine}} = \mathcal{N}_\phi(M^{2D}_{\text{coarse}}, \mathbf{I}_t)
\end{equation}
where $\mathcal{S}$ denotes the semantic field obtained from Stage I, $v_t$ represents the camera viewpoint, $\mathcal{R}(\cdot)$ signifies the Gaussian Splatting rasterizer, and $\mathcal{G}$ is the Gaussian model. $\mathbf{I}_t$ corresponds to the rendered RGB image at viewpoint $v_t$. $M^{2D}_{\text{refine}}$ corresponds to the refined 2D mask.

\subsubsection{Network Architecture}
As illustrated in Fig.~\ref{fig:pipeline}, the refinement network employs a dual-branch encoder-decoder architecture specifically designed to fuse appearance cues with coarse semantic guidance.

\noindent\textbf{Dual-Branch Encoder:} The network processes two inputs: the rendered RGB image and the coarse rasterized mask.
\begin{itemize}
    \item \textit{Image Encoder:} We employ a ResNet-50 backbone initialized with ImageNet pre-trained weights to extract multi-scale appearance features. We strategically select this robust architecture to leverage its rich feature hierarchy, which significantly accelerates the convergence of our online distillation process compared to training a lightweight encoder from scratch. This branch captures high-fidelity texture and boundary details essential for completing the coarse mask.

    \item \textit{Mask Encoder:} A lightweight convolutional branch processes the coarse mask $M^{2D}_{\text{coarse}}$. This serves as a strong spatial prompt, guiding the network to focus on specific regions of interest.
\end{itemize}
\noindent\textbf{Feature Fusion and Decoding:} 
High-level semantic features from the Mask Encoder are fused with deep appearance features from the Image Encoder. To effectively integrate these modalities, we employ a Channel Attention (CA) mechanism before concatenation. The decoder follows a U-Net-like structure with skip connections, progressively upsampling features to recover fine-grained details.

\subsubsection{Online Distillation Strategy}
Unlike traditional supervised learning requiring manual annotation, we adopt an online distillation strategy. We leverage the temporally consistent masks generated by the foundation model on the rendered views as \textit{pseudo-ground truth}. Since the network focuses on the simplified task of refining boundaries and filling holes based on specific scene textures, rather than learning generalized semantics, it converges rapidly. Empirically, this adaptation requires approximately 4 minutes of fine-tuning per scene.

Formally, the network is optimized via a pixel-wise multi-class cross-entropy loss over the training views $\mathcal{V}$:
\begin{equation}
    \mathcal{L}_{\text{refine}} = -\frac{1}{|\mathcal{V}| \cdot |\Omega|} \sum_{v \in \mathcal{V}} \sum_{p \in \Omega} \sum_{c=1}^{C} \mathbb{1}[M_{v}^{\text{GT}}(p)=c] \log \hat{M}_{v}(p, c)
\end{equation}
where $\Omega$ denotes the pixel coordinates space of the image, and $C$ represents the total number of semantic classes. The term $\mathbb{1}[\cdot]$ is the indicator function which equals 1 if the condition is true and 0 otherwise. Specifically, $M_{v}^{\text{GT}}(p)$ denotes the pseudo-ground truth label index for pixel $p$ in view $v$, while $\hat{M}_{v}(p, c)$ represents the predicted probability that pixel $p$ belongs to class $c$.

Once trained, this module replaces the foundation model entirely, enabling high-speed inference (approx. 27 ms/frame) for arbitrary, non-continuous viewpoints while maintaining high segmentation quality.

\section{Experiments}
\subsection{Datasets}
Our evaluation utilizes two primary datasets: NVOS~\cite{nvos} and the SPIn-NeRF benchmark~\cite{spinerf}. The NVOS dataset extends the LLFF dataset~\cite{llff} by providing instance-level annotations for foreground objects across diverse scenes. The SPIn-NeRF benchmark is a multiview segmentation dataset that offers unified instance segmentation labels across multiple domains. Both datasets are specifically designed for NeRF-based and Gaussian-based segmentation tasks, ensuring direct compatibility with our methods.

To comprehensively evaluate cross-domain performance, we conduct additional experiments on the KITTI-360~\cite{kitti360}, UDD~\cite{UDDdataset}, NeRDS-360~\cite{nerds360}, Mip-360~\cite{mip360}, LERF-Mask dataset~\cite{gaussian_grouping}, DesktopObjects-360~\cite{pointgauss} scenes. This extended evaluation enables a systematic analysis of our method's segmentation quality (in both 2D and 3D) and computational efficiency across: 1) indoor/outdoor environments, 2) varying object densities, and 3) different illumination conditions.

\subsection{Experimental Settings}
\label{sec:settings}
All experiments were conducted on a workstation equipped with an NVIDIA RTX 4090 GPU and an Intel i5-11400F CPU, running PyTorch~2.1.0 with CUDA~11.8 on a Linux operating system. Unless otherwise specified, all Gaussian scenes used for evaluation were trained for 30,000 iterations. Because SAM2 requires manually provided prompts to segment the target of interest, we simulate this behavior during initialization by randomly sampling clicks on the ground-truth object to form the prompt. No additional user inputs are required throughout inference.

For datasets with sequential image organization, we adopt a filename-based pseudo-temporal ordering to facilitate initialization. Specifically, for the Spin-NeRF dataset, the training images are lexicographically sorted by filename to form the pseudo-temporal sequence. Likewise, for the NVOS dataset, we establish the same ordering strategy and initialize the target region through random clicks on the ground-truth object.

\subsection{Quantitative Segmentation Performance}
\label{sec:accuracy}
We evaluate our method on two challenging benchmarks: the NVOS dataset~\cite{nvos} and the SPIn-NeRF dataset~\cite{spinerf}. We compare SAGOnline against a comprehensive set of baselines, including NVOS~\cite{nvos}, ISRF~\cite{isrf}, MVSeg~\cite{spinerf}, SA3D~\cite{sa3d}, OmniSeg3D-GS~\cite{omniseg}, SA3D-GS~\cite{sa3d-gs}, SAGA~\cite{SAGAgaussian}, and Click-Gaussian~\cite{ClickGaussians}. Notably, OmniSeg3D-GS represents the previous top performer on SPIn-NeRF, while SAGA holds the leading position on NVOS.

\begin{table}[t]
\centering
\caption{Quantitative comparison on NVOS and SPIn-NeRF datasets. Our SAGOnline achieves state-of-the-art performance on both benchmarks. ``-'' indicates the method is not applicable or not reported for that dataset.}
\label{tab:combined_results}
\small
\resizebox{\linewidth}{!}{
\begin{tabular}{l cc c cc}
\toprule
\multirow{2}{*}{\textbf{Method}} & \multicolumn{2}{c}{\textbf{NVOS Dataset}} & & \multicolumn{2}{c}{\textbf{SPIn-NeRF Dataset}} \\
\cmidrule{2-3} \cmidrule{5-6}
& mIoU & mAcc & & mIoU & mAcc \\
\midrule
NVOS~\cite{nvos} & 70.1 & 92.0 & & - & - \\
ISRF~\cite{isrf} & 83.8 & 96.4 & & - & - \\
MVSeg~\cite{spinerf} & - & - & & 91.0 & 98.9 \\
SA3D~\cite{sa3d} & 90.3 & 98.2 & & 92.4 & 98.9 \\
OmniSeg3D~\cite{omniseg} & 91.7 & 98.4 & & \textbf{95.2} & 99.2 \\
SA3D-GS~\cite{sa3d-gs} & 92.2 & 98.5 & & 93.2 & 99.1 \\
SAGA~\cite{SAGAgaussian} & 92.6 & 98.6 & & 93.4 & 99.2 \\
Click-Gaussian~\cite{ClickGaussians} & - & - & & 94.0 & - \\
\midrule
\textbf{SAGOnline (Ours)} & \textbf{92.7} & \textbf{98.7} & & \textbf{95.2} & \textbf{99.3} \\
\bottomrule
\end{tabular}
}
\end{table}

\textbf{Performance on NVOS.} As presented in Table~\ref{tab:combined_results}, SAGOnline achieves a mIoU of 92.7\% and mAcc of 98.7\%, effectively matching and slightly surpassing the previous state-of-the-art method, SAGA. 
It is crucial to note that SAGA relies on a heavier offline optimization process. The fact that SAGOnline achieves a slight edge (+0.1\% mIoU) confirms that our streamlined online framework does not compromise segmentation quality.

\textbf{Performance on SPIn-NeRF.} Table~\ref{tab:combined_results} shows the results on the SPIn-NeRF dataset. Our method attains 95.2\% mIoU and 99.3\% mAcc. While the mIoU is tied with the previous best performer (OmniSeg3D), our method shows a slight improvement in segmentation consistency (mAcc). 
More importantly, these results indicate that SAGOnline effectively utilizes the priors from foundation models.

\textbf{Analysis of Performance Saturation.}
As observed in Table~\ref{tab:combined_results}, the performance gap between recent methods (e.g., SA3D-GS, SAGA, OmniSeg3D) and ours is relatively narrow. This convergence suggests that current 3D segmentation methods are approaching the upper bound defined by the quality of the underlying 2D foundation models (e.g., SAM series). In this context, SAGOnline achieves high accuracy within an efficient online framework. It bridges 2D priors and 3D Gaussians without the heavy computational overhead of prior offline methods.

Unlike baselines that require extensive offline training or complex distillation to preserve foundation model priors, our method leverages the temporal coherence of SAM 2 to directly and rapidly propagate segmentation in 3D. The results demonstrate that SAGOnline effectively bridges the gap between 2D foundation models and 3D Gaussian Splatting, achieving state-of-the-art accuracy without the computational overhead of prior methods.

\begin{figure*}[t]
    \centering
    \includegraphics[width=1.0\linewidth]{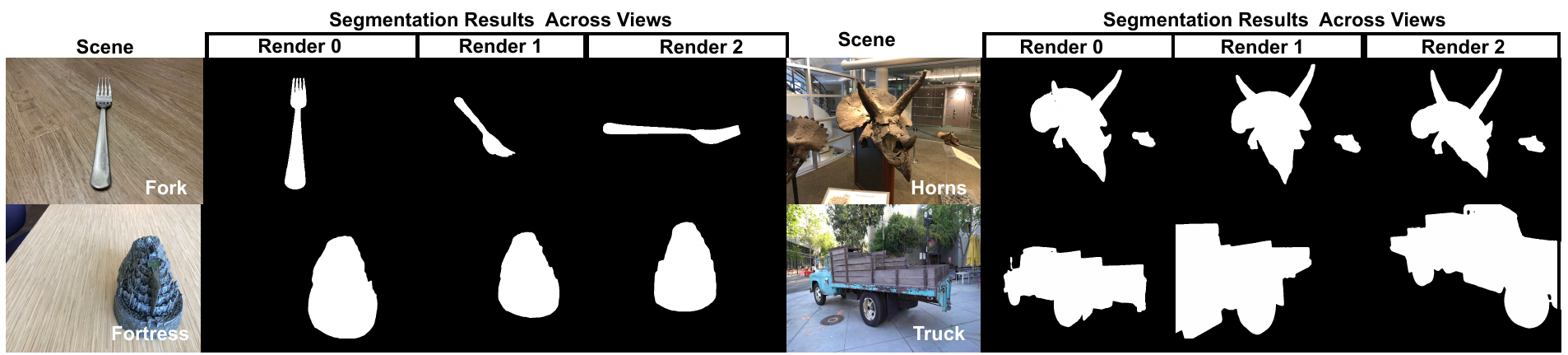}
    \caption{Qualitative segmentation results demonstrating multi-view consistency. We present target object extraction results for four diverse scenes: Fork, Fortress, Horns, and Truck. For each block, the input scene is shown in the 1st column, followed by the extracted binary masks rendered from three distinct viewpoints (Render 0-2). The results highlight our method's ability to maintain precise object boundaries and geometric coherence across varying camera poses.}
    \label{fig:results}
\end{figure*}

Qualitative results are illustrated in Fig.~\ref{fig:results}. Our method exhibits robust performance across diverse scenarios. For instance, in the Fork scene, SAGOnline captures fine structural details that are occasionally missed by ground truth annotations, further validating the effectiveness of our initialization and refinement strategy.

\subsection{Time Efficiency}
\label{sec:time_efficiency}

The temporal efficiency of 3D segmentation algorithms is a critical factor for their deployment in latency-sensitive applications, such as AR/VR systems and interactive scene editing. We evaluate the efficiency of SAGOnline against state-of-the-art baselines on DesktopObjects-360 dataset, focusing on two key metrics:
\begin{itemize}
    \item \textbf{Time-to-First-Mask (Start-up Latency):} The waiting time required before the user can visualize and interact with the first valid segmentation result.
    \item \textbf{Inference Speed:} The rendering frame rate during the interactive segmentation session.
\end{itemize}

To ensure a fair comparison, all baseline methods were executed on the same hardware configuration (NVIDIA RTX 4090). The quantitative comparisons are summarized in Table~\ref{tab:time_efficiency}.

\begin{table}[t]
\caption{Time Efficiency Comparison on Multi-object Scenes. Unlike offline methods that require significant pre-computation (Blocking), our method leverages a Foundation Model to offer instant interaction (0 s), with rapid 3D mask aggregation (1.47s for $\sim$200 frames).}
\label{tab:time_efficiency}
\centering
\resizebox{\linewidth}{!}{
\begin{tabular}{lccc}
\toprule
\textbf{Method} & \textbf{Time-to-First-Mask} & \textbf{Inference} & \textbf{Mode} \\ 
& \textit{(Start-up Latency)} & \textit{(Speed)} & \\
\midrule
SA3D~\cite{sa3d} & \textbf{0 s} & $>$50 s & Online \\
GARField~\cite{garfield} & 45 min & 3 s & Offline (Blocking) \\
Feature3DGS~\cite{feature3dgs} & 35 min & 510 ms & Offline (Blocking) \\
OmniSeg3D-GS~\cite{omniseg} & 37 min & 463 ms & Offline (Blocking) \\
SAGA~\cite{SAGAgaussian} & 32 min & 31 ms & Offline (Blocking) \\
\midrule
\textbf{SAGOnline (Initial)} & \textbf{0 s}$^\dagger$ & 97 ms & \textbf{Online (Immediate)} \\
\textbf{SAGOnline (Refined)} & \textit{$\sim$5 min$^*$} & \textbf{27 ms} & \textbf{Online (Background)} \\
\bottomrule
\end{tabular}
}
\vspace{2pt}
\footnotesize{
\\ \textbf{$^\dagger$} Instant response via Foundation Model. Full 3D aggregation for the sequence takes approx. 1.47s. 
\\ \textbf{$^*$} The refinement process runs asynchronously in the background. Users can interact immediately using the \textit{Initial} mode without waiting.}
\end{table}
\textbf{Immediate Availability vs. Blocking Pre-computation.} 
As shown in Table~\ref{tab:time_efficiency}, traditional offline methods (e.g., GARField, SAGA, Feature3DGS) suffer from a significant "cold-start" problem. These approaches are \textit{blocking}: users must wait for a mandatory training period of 30 to 45 minutes before any segmentation result is produced. 

In contrast, SAGOnline is designed as a \textit{non-blocking}, online framework. By leveraging a vision Foundation Model (e.g., SAM2), our method instantly generates a 2D mask, achieving a \textbf{Time-to-First-Mask of 0 s}. Following this interaction, the system rapidly aggregates the generated 2D masks (e.g., across 207 frames) to construct the initial 3D mask. This aggregation process completes in merely \textbf{1.47 seconds}, which is negligible compared to the tens of minutes of training required by offline baselines. This capability matches the responsiveness of online methods like SA3D but operates at a significantly higher frame rate (97 ms vs. $>50$s).

\textbf{Asynchronous Self-supervised Refinement.}
To achieve real-time rendering performance suitable for AR/VR (e.g., $>30$ FPS), our framework incorporates a Self-supervised Refinement Module. This module operates as a background process to distill the coarse 3D masks and optimize Gaussian attributes, which takes approximately 5 minutes.
Crucially, this refinement phase is \textbf{transparent to the user}. The user can continue to interact with the scene using the \textit{Initial} mode (97 ms/frame) while the optimization runs in parallel. Once the refinement converges, the system seamlessly transitions to the \textit{Refined} mode, boosting the inference speed to \textbf{27 ms/frame} (37 FPS)—a $13\%$ improvement over SAGA and $17\times$ faster than OmniSeg3D-GS.

This "As-You-Go" paradigm ensures that SAGOnline provides the best of both worlds: the instant responsiveness of online methods and the high-performance rendering of offline baked models, without imposing mandatory waiting periods on the user.

\begin{figure*}[t]
    \centering
    \includegraphics[width=1.0\linewidth]{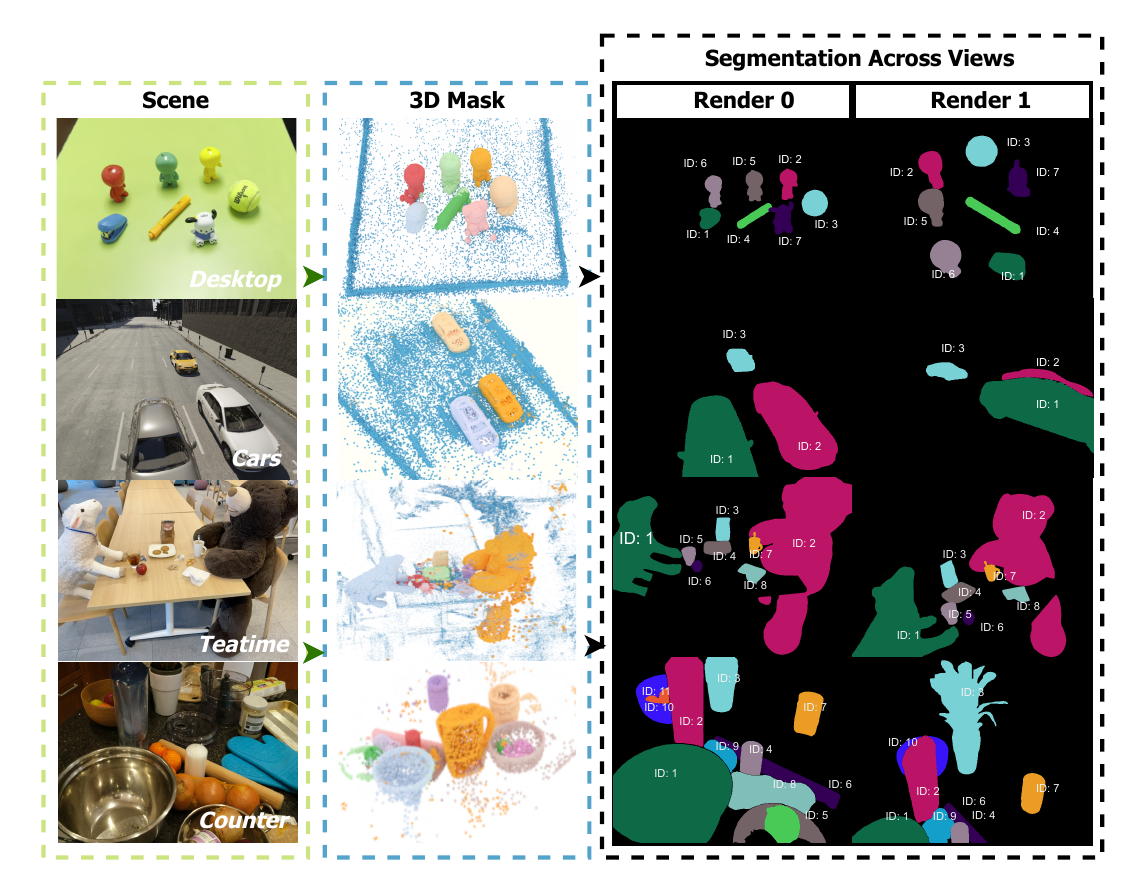}
    \caption{Qualitative results of multi-object segmentation across diverse scenes. The figure is organized into four scene blocks: Desktop, Cars, Teatime, and Counter. For each block, the 1st column displays the reference scene; the 2nd column visualizes the explicit 3D masks formed by the Gaussian primitive means; and the subsequent columns (3rd-4th) present the segmentation masks rendered from different viewpoints. Consistent colors across views indicate that our method maintains robust instance identity.}
    \label{fig:multi_object}
\end{figure*}

\subsection{Multi-Object Segmentation \& Cross-View Consistency}
While current benchmarks such as NVOS and Spin-NeRF provide valuable evaluation platforms for 3D-aware segmentation, they exhibit two major limitations: (1) scenes in these datasets generally contain only a single dominant object of interest, limiting the evaluation of instance-level separation in cluttered environments; and (2) their evaluation protocols primarily rely on 2D projection-based metrics (e.g., IoU and mIoU on rendered masks), which cannot comprehensively capture true 3D spatial consistency or inter-object boundary precision. To overcome these limitations and comprehensively assess the generalization capability of our method, we additionally evaluate on the Mip-360, LERF-Mask, DesktopObjects-360, and NeRds360 datasets. These datasets feature complex indoor, tabletop, and outdoor scenes characterized by multiple interacting instances, severe occlusions, and varying object scales.

Fig.~\ref{fig:multi_object} presents representative qualitative results of our method across these datasets. As shown in the second column, our approach successfully distinguishes individual object instances from a 3D perspective, achieving accurate separation even in scenes with severe occlusion and background clutter. The third and fourth columns further demonstrate consistent segmentation across drastically different viewpoints. Although the scenes are static, the significant camera movement creates a challenging scenario akin to object tracking. In this context, our method ensures that each object maintains a stable instance ID throughout the trajectory. This multi-view consistency highlights our framework’s capacity for reliable cross-view instance association and tracking-like continuity.

The robustness of our framework stems from the first-stage generation of global 3D instance masks, which establish explicit 3D coordinate correspondences for every object. These masks serve as spatial anchors that enable stable object localization and identity preservation under arbitrary viewpoint transformations. Leveraging the Gaussian Splatting rendering pipeline, our system projects 3D segmentation results onto the 2D image plane while estimating per-pixel opacity. This enables accurate reasoning about inter-object occlusion and visibility, allowing the method to maintain object continuity even in cluttered scenes. Consequently, our method achieves high-fidelity multi-object segmentation with temporal-like consistency across views, offering a practical foundation for downstream applications such as 3D scene editing, object-centric reconstruction, and Gaussian field compositing.

\subsection{Different Segmentation Tasks}
\begin{figure*}[t]
    \centering
    \includegraphics[width=1.0\linewidth]{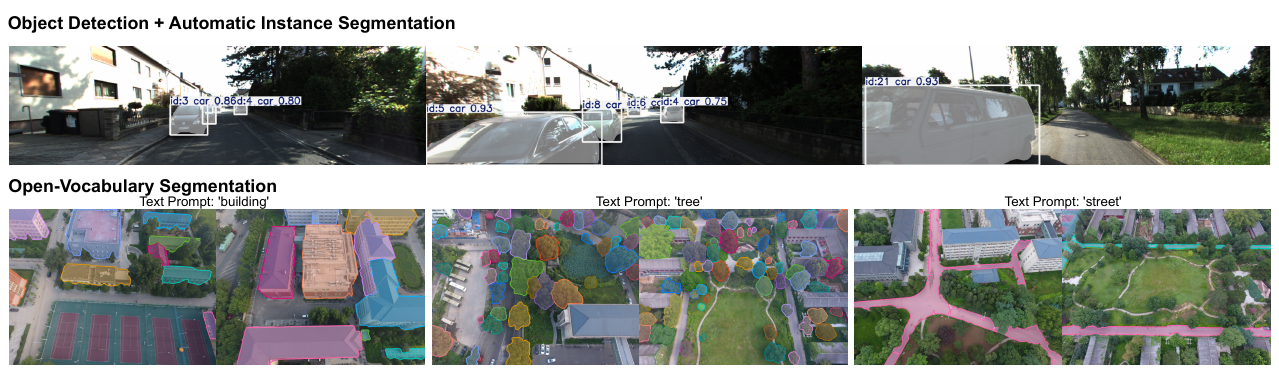}
    \caption{Qualitative results on diverse segmentation tasks. The top row demonstrates automatic vehicle instance segmentation on the KITTI-360 dataset utilizing YOLO. The bottom row showcases open-vocabulary segmentation on the UDD dataset driven by SAM 3 with text prompts. These results highlight our framework's versatility, capable of leveraging distinct backbone models to handle various segmentation paradigms within 3D Gaussian Splatting scenes.}
    \label{fig:variousTask}
\end{figure*}

Our framework enables seamless integration with various foundation models, allowing it to support multiple segmentation tasks across different domains. This design also ensures that our method remains compatible with future algorithmic upgrades. For example, by incorporating YOLO~\cite{yolo11_ultralytics}, we can perform fast object detection and automatic instance segmentation. We evaluate this capability on the widely used autonomous-driving dataset KITTI-360, as illustrated in Fig.~\ref{fig:variousTask}. Our method effectively leverages YOLO to accomplish instance segmentation of vehicles.

Moreover, our framework naturally supports language-guided segmentation tasks through direct integration with vision-language models. For instance, by using the recent Segment Anything Model 3 (SAM~3)~\cite{sam3}, we can achieve open-vocabulary segmentation. We conduct experiments on the UAV-based UDD dataset, and the visualizations show that our method can accurately segment buildings, roads, and vegetation in complex urban scenes.

\begin{figure}[t]
    \centering
    \includegraphics[width=1.0\linewidth]{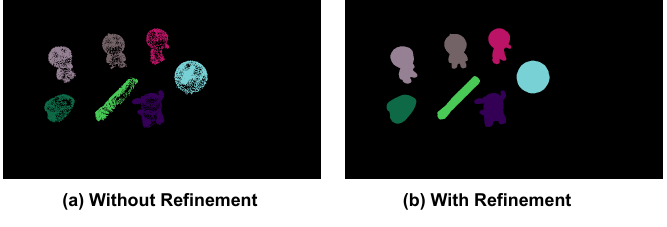}
    \caption{Visual comparison of the Dual-Branch Refinement Network. (a) Segmentation results derived directly from Sparse Semantic 3D Gaussians exhibit noticeable sparsity and noise artifacts. (b) After applying our refinement module, the segmentation masks become significantly denser, smoother, and spatially coherent, demonstrating the network's effectiveness in enhancing mask quality.}
    \label{fig:postprocessing}
\end{figure}
\subsection{Effectiveness of Refinement Module}
\label{sec:ablation}
The Refinement algorithm is designed to refine the rendered segmentation masks into spatially coherent and accurate results. To verify the effectiveness of our refinement, we perform experiments comparing results before and after the Refinement Module.

As illustrated in Fig.~\ref{fig:postprocessing}, the directly rendered masks contain scattered and fragmented regions. After the Refinement Module, these sparse points are aggregated into continuous and well-defined instance masks. This refinement significantly enhances both mask completeness and boundary precision, yielding more stable multi-view consistent segmentation results.

\subsection{Robustness Analysis}
\label{suppl:robust}
Since our algorithm reconstructs 3D masks from multi-view 2D segmentation results, it is essential to examine how the number of available 2D inputs affects the quality and stability of the generated 3D masks. To this end, we conducted a robustness analysis by randomly sampling different numbers of 2D masks as input for the 3D mask generation process. As the number of input masks increased, we measured both the quality of the reconstructed 3D masks and the corresponding computation time. This experiment provides a quantitative understanding of the trade-off between accuracy and efficiency, and offers insights into the scalability of our method under varying data availability.

\begin{figure*}
    \centering
    \includegraphics[width=\linewidth]{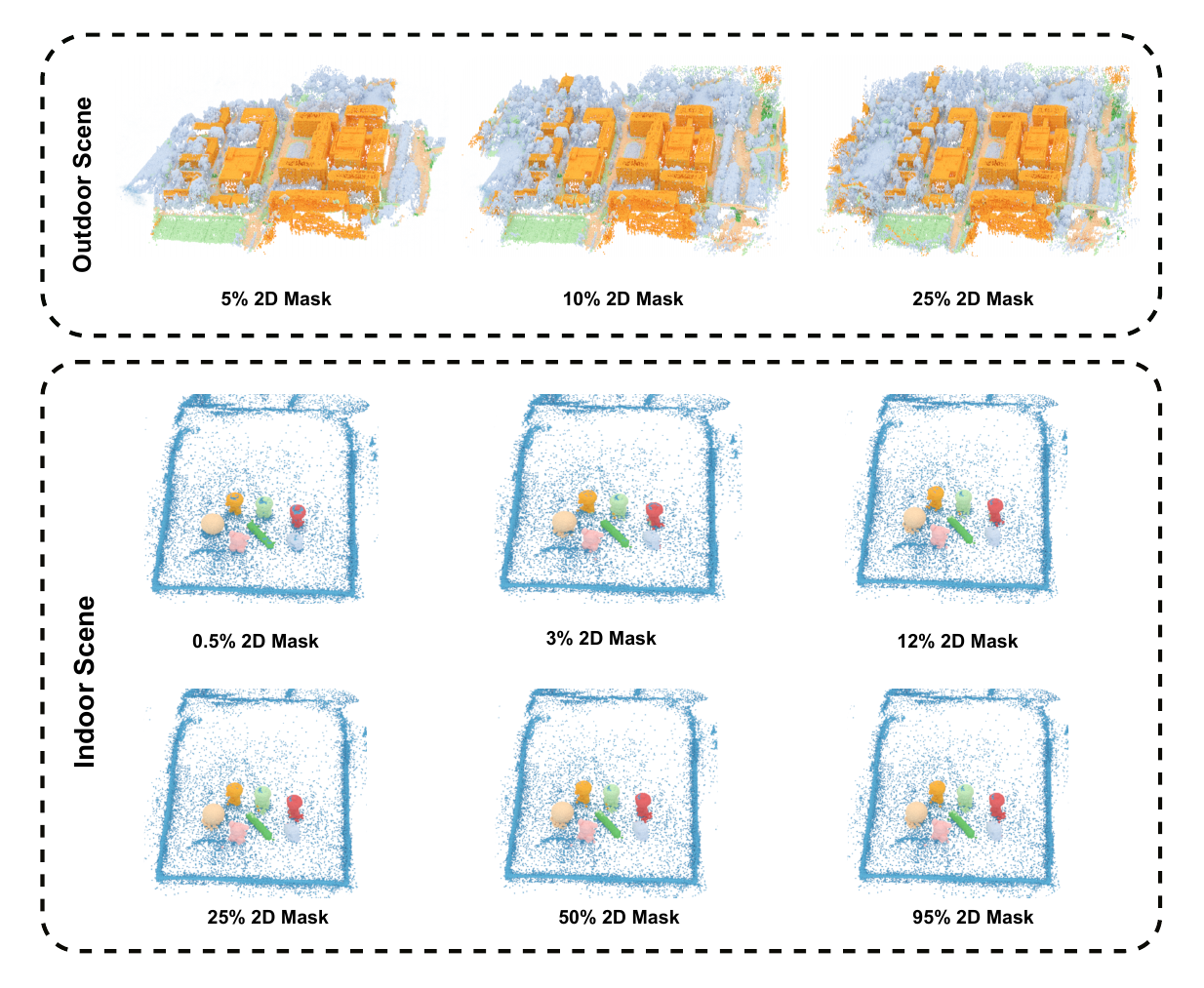}
    \caption{Qualitative evaluation of 3D segmentation robustness under varying 2D mask supervision. The top row illustrates the outdoor scene results with 2D mask usage ranging from 5\% to 25\%, while the bottom row depicts the indoor scene from 0.5\% to 95\%. Observe that our method maintains consistent semantic structures even with extremely sparse 2D mask inputs (e.g., 5\% in the outdoor scene).}
    \label{fig:robust}
\end{figure*}

As illustrated in Fig.~\ref{fig:robust}, when only a few 2D masks are provided, the resulting 3D masks still preserve clear object boundaries and instance separability. As more 2D views are incorporated, the spatial completeness and surface smoothness of the 3D masks further improve, with more Gaussian primitives being assigned consistent instance IDs. Remarkably, even with as few as six input masks, our algorithm is capable of accurately reconstructing multiple objects with minimal identity confusion, indicating strong robustness against sparse input views.

In terms of efficiency, Table~\ref{tab:robust} reports the average time required to generate 3D masks under different input settings. The results show a near-linear relationship between the number of input masks and computation time: the full 206-view configuration requires 1466~ms, while processing only a single mask reduces the time to 111~ms. This demonstrates that our approach achieves efficient and scalable 3D mask reconstruction, maintaining real-time performance even under limited input conditions. The combination of high robustness and low latency makes the method suitable for practical multi-view perception and online scene understanding tasks.

\begin{table}[t]
\centering
\caption{Time consumption of 3D mask generation. "Avg." denotes the average time cost per mask (ms).}
\label{tab:robust}
\setlength{\tabcolsep}{2.5pt} 
\begin{tabular}{lcc|lcc}
\toprule
\# Masks & Time (ms) & Avg. (ms) & \# Masks & Time (ms) & Avg. (ms) \\
\midrule
Full (206) & 1466 & 7.1 & 12 & 162 & 13.5 \\
100        & 571  & 5.7 & 6  & 140 & 23.3 \\
50         & 384  & 7.7 & 3  & 114 & 38.0 \\
25         & 224  & 9.0 & 1  & 111 & 111.0 \\
\bottomrule
\end{tabular}
\end{table}

\subsection{Fidelity Preservation in Refinement Stage}
\label{suppl:stage2}
To evaluate the impact of the Refinement Module on segmentation fidelity, we analyzed the performance shift between the \textit{Initial} and \textit{Refined} stages. The Refined stage maintains an overall accuracy of 99.89\% and mIoU of 98.47\% relative to the Initial stage outputs. This high correlation confirms that our background refinement process successfully optimizes the Gaussian attributes for rendering speed (from 97ms to 27ms) without degrading the geometric integrity or semantic accuracy established in the initialization phase.

\section{Conclusion}
\label{sec:conclusion}
In this work, we introduced SAGOnline, an optimization-free framework that enables real-time, multi-view-consistent segmentation for 3D Gaussian Splatting scenes. Departing from prior approaches that rely on heavy feature distillation, continuous semantic embeddings, or lengthy offline optimization, SAGOnline reformulates Gaussian segmentation as three lightweight sub-tasks: novel-view rendering, multi-view 2D segmentation, and global 3D fusion. Central to our framework is the Rasterization-aware Geometric Consensus mechanism. It exploits 3DGS rasterizer traceability to deterministically lift 2D predictions into explicit, discrete 3D semantics. This explicit formulation allows immediate mask extraction from any viewpoint without feature decoding, while our self-supervised refinement module restores fine-grained boundaries and enables high-speed rendering suitable for interactive applications.

Extensive experiments across NVOS, SPIn-NeRF, KITTI-360, UDD, DesktopObjects-360, and other datasets demonstrate that SAGOnline achieves state-of-the-art segmentation performance while offering the fastest initialization and inference speeds among existing 3DGS-based methods. The framework supports diverse segmentation paradigms—instance, semantic, and prompt-guided segmentation—through a modular interface that integrates a wide range of foundation models, including SAM2, SAM3, YOLO, and video segmenters. These results show that SAGOnline not only preserves the accuracy of powerful 2D foundation models in the 3D domain, but also delivers the responsiveness required for real-time scene understanding, interactive 3D editing, and immersive AR/VR applications.

Overall, SAGOnline demonstrates that explicit, geometry-aware fusion combined with foundation-model priors provides a practical and scalable route toward high-fidelity 3D segmentation in Gaussian-splatting–based representations, paving the way for more general and interactive 3D perception systems.

\bibliographystyle{IEEEtran}
\bibliography{ref}


 





\end{document}